\documentclass[10pt,twocolumn,letterpaper]{article}

\usepackage{cvpr}              

\usepackage{times}
\usepackage{epsfig}
\usepackage{graphicx}
\usepackage{amsmath}
\usepackage{amssymb}

\definecolor{cvprblue}{rgb}{0.21,0.49,0.74}
\usepackage[pagebackref,breaklinks,colorlinks,allcolors=cvprblue]{hyperref}

\def\paperID{17} 
\def\confName{CVPR}
\def\confYear{2025}

\definecolor{grayhighlight}{RGB}{213,229,255}

\newsavebox\CBox

\begin{document}
	
	\title{Learned Smartphone ISP on Mobile GPUs: \\ Mobile AI 2026 Challenge Report}
	
	\author{
		Andrey Ignatov \and Georgy Perevozchikov \and Radu Timofte 
		\and Yuhong He \and Yuxuan Qiu \and Tianyu Hao \and Ruoqi Li \and Yueqi Yang \and Chaoyu Feng \and Dongqing Zou \and Lei Lei \and Na Jiang
		\and Yongrui Zhang \and Hailong Yan \and Xiangtao Zhang \and Ao Li \and Le Zhang
		\and Andrei Arhire
		\and Jing Fang \and XingLin Xie \and YiChen Lu
		\and Haibo Xu
	}
	
	\maketitle

	\begin{abstract}
		 
		  With the increasing importance of mobile camera quality in modern smartphones, growing attention has been directed toward computational photography and advanced image signal processing (ISP) algorithms aimed at enhancing various perceptual attributes of captured images.
		  This creates a need for efficient deep learning models for this problem that are not only performant but can also be deployed on constrained mobile hardware. In this Mobile AI challenge, we address this problem and propose the participants to design fast solutions that can reconstruct RGB photos from RAW Bayer camera data while running on popular smartphone GPUs from Qualcomm and Arm. For this, the participants were provided with a large-scale Fujifilm UltraISP dataset consisting of RAW-RGB image
		  pairs captured with the Sony IMX586 Quad Bayer mobile sensor and a professional 102MP medium format FujiFilm GFX100 camera. The runtime of all models was evaluated on the Snapdragon 8 Elite Gen 5 and Dimensity 9500 chipsets with the Adreno and Mali GPUs, respectively. Top solutions proposed in this challenge are able to reconstruct high-quality 5MP RGB image from raw data on both platforms under 100 milliseconds. A comprehensive description of the models developed in the challenge is provided in this paper.
		
	\end{abstract}
	{\let\thefootnote\relax\footnotetext{%
			\hspace{-5mm}$^*$Andrey Ignatov, Georgy Perevozchikov and Radu Timofte are the organizers of the Mobile AI 2026 challenge \textit{(andrey@vision.ee.ethz.ch, georgii.perevozchikov@uni-wuerzburg.de, radu.timofte@uni-wuerzburg. de)}. The other authors participated in the challenge. \\ Appendix \ref{sec:apd:team} contains the authors' team names and affiliations.\\ \\ Mobile AI 2026 Workshop website: \\ \url{https://ai-benchmark.com/workshops/mai/2026/}
	}}
	
	\section{Introduction}
	
\begin{figure*}[t!]
	\centering
	\setlength{\tabcolsep}{1pt}
	\resizebox{\linewidth}{!}
	{
		\begin{tabular}{cccc}
			\scriptsize{Sony IMX586 RAW -- Visualized}\normalsize & \scriptsize{Sony IMX586 -- MediaTek ISP}\normalsize & \scriptsize{Fujifilm Camera}\normalsize\\
			\includegraphics[width=0.33\linewidth]{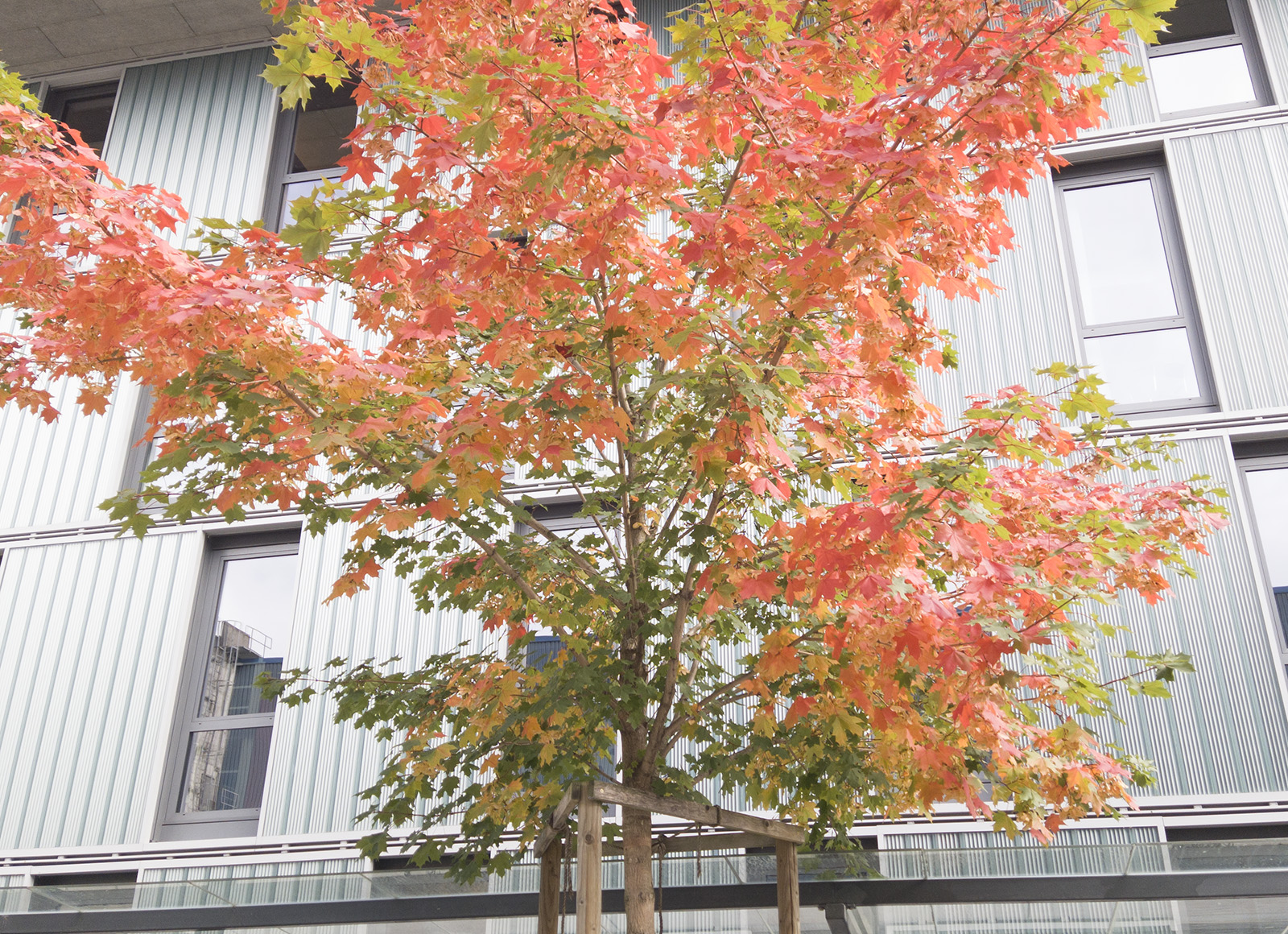}&
			\includegraphics[width=0.33\linewidth]{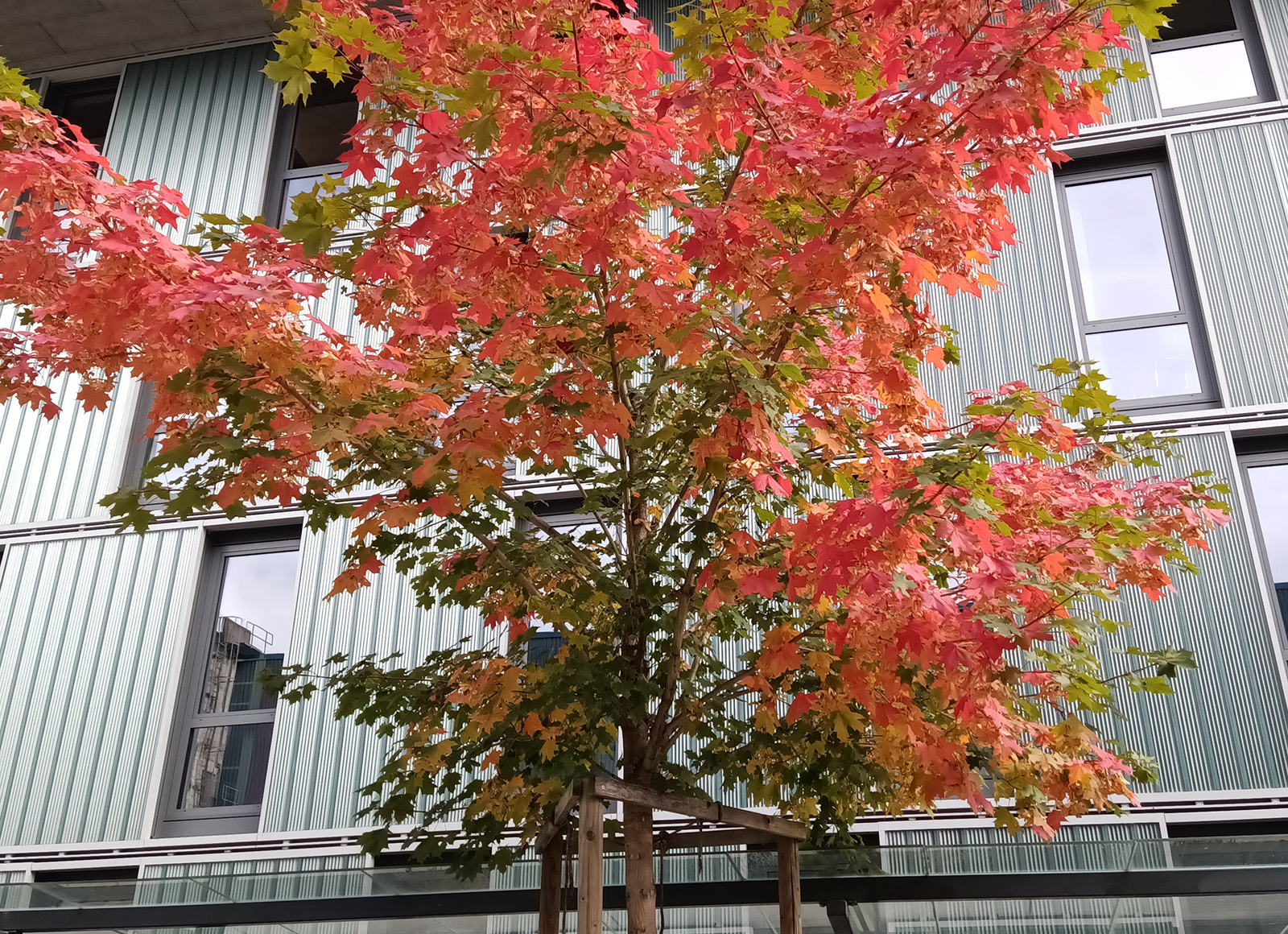}&
			\includegraphics[width=0.33\linewidth]{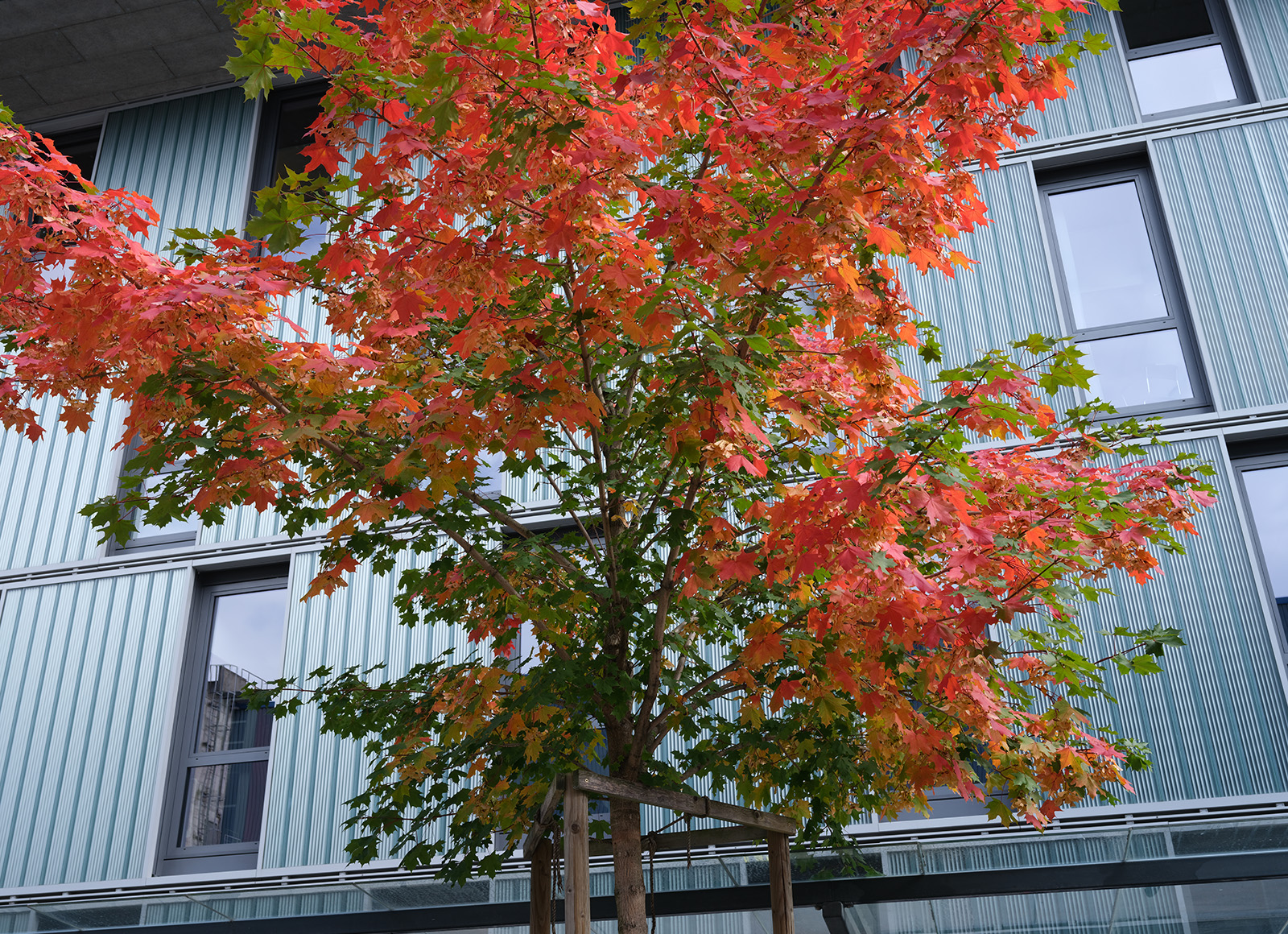}
		\end{tabular}
	}
	\vspace{0.2cm}
	\caption{A sample set of images from the collected dataset.  From left to right: the original RAW image visualized with Photoshop's built-in raw image processing engine, RGB image obtained with MediaTek's built-in ISP system, and the target Fujifilm photo.}
	\label{fig:example_photos}
\end{figure*}
	
	A primary driver of performance improvements in smartphone cameras can be attributed to sophisticated image processing algorithms, including those for noise suppression, accurate color reproduction, and high dynamic range (HDR) imaging. While individual stages of the traditional image signal processing (ISP) pipeline can be replaced by dedicated deep learning models to enhance reconstruction quality, a more comprehensive paradigm involves substituting the entire RAW-to-RGB pipeline with a unified neural network that implicitly learns all intermediate processing steps. Such an approach enables direct utilization of the richer 10--16 bit information contained in the original, unprocessed RAW Bayer data, thereby potentially leading to improved image restoration performance. In~\cite{ignatov2020replacing}, the authors first introduced this approach by constructing the Zurich RAW-to-RGB dataset comprised of paired RAW and RGB images captured using both a mobile camera sensor and a high-end DSLR camera. This dataset was subsequently used to train a deep neural network for end-to-end image reconstruction. The experimental results demonstrated the strong potential of this method in comparison to the commercial hand-crafted ISP of the Huawei P20 smartphone. Even better results were achieved by the models proposed in the follow-up challenges and papers using the same dataset~\cite{ignatov2020aim,dai2020awnet,silva2020deep,kim2020pynet,ignatov2019aim}. A further advancement in deep learning–based ISP development was presented in~\cite{ignatov2022microisp,ignatov2022pynetv2}, where the authors introduced a more advanced Fujifilm UltraISP dataset for the considered task along with mobile-efficient CNN architectures capable of processing high-resolution images directly on smartphones. In this competition, we use the same dataset and propose the participants to design even more efficient learned ISP models that can run on recent smartphone GPUs.
		
	Designing deep learning models that run efficiently on mobile devices~--- especially across different AI accelerators like NPUs~--- is a complex challenge. It demands careful analysis of hardware constraints such as limited operator support, restricted memory capacity and bandwidth, reduced accuracy for quantized models and specific driver behavior of NPUs. Even when certain layers or operations are technically supported, they may be executed inefficiently due to constraints related to NPU hardware architecture or its driver implementation, and thus create performance bottlenecks during inference. A striking example here would be the Google Tensor NPU present in all recent Google Pixel smartphones that, despite having quite high theoretical peak performance, is unable to efficiently accelerate the vast majority of even simplistic CNN-based models~\cite{ignatov2025MAISR}. As a result, achieving low latency and energy efficiency often requires partial or complete redesigns of model architectures. For a more detailed discussion of these challenges and the performance of modern mobile AI accelerators we refer to~\cite{ignatov2019ai,ignatov2018ai}. Overall, all these limitations require various adaptations of the standard deep learning models for their efficient deployment on mobile NPUs, including conventional optimizations like 16-bit / 8-bit~\cite{chiang2020deploying,jain2019trained,jacob2018quantization,yang2019quantization,li2021mqbench,nagel2021white,bondarenko2021understanding,siddegowda2022neural,tan2024mobilequant,lin2025awq} and low-bit~\cite{cai2020zeroq,uhlich2019mixed,banner2019post,ignatov2020controlling,liu2018bi,zhuo2022empirical,bhalgat2020lsq+,kim2023memory,dettmers2023case,bondarenko2024low,li2025memory} quantization, network pruning and compression~\cite{chiang2020deploying,li2019learning,
		liu2019metapruning,obukhov2020t,kim2021pqk,hawks2021ps,zheng2022model,xu2023sparsebytenn,kuzmin2023pruning,wang2024towards}, NPU-specific adaptations, platform-aware neural architecture search~\cite{howard2019searching,tan2019mnasnet,wu2019fbnet,
		wan2020fbnetv2,mills2021profiling,chu2021discovering,zhan2021achieving,wu2022compiler,chitty2022neural,yoshihama2023hardware}, \etc.
	
	In this Mobile AI Challenge, the participants were proposed to develop efficient deep learning models that can reconstruct 5MP RGB image from RAW Bayer data while running on smartphone GPUs. For this, the participating teams were provided with the Fujifilm UltraISP dataset~\cite{ignatov2022microisp} consisting of thousands paired photos captured with a normal mobile camera sensor and a professional 102MP medium-format FujiFilm GFX100 camera. The runtime of the models was evaluated on mobile GPUs present in the Snapdragon 8 Elite Gen 5 and the Dimensity 9500 chipsets. Each submission was evaluated based on both image fidelity and runtime efficiency, thus balancing reconstruction quality with model efficiency. All solutions developed during this challenge are fully compatible with the LiteRT~\cite{LiteRT2025} (former TensorFlow Lite) framework, enabling deployment on a variety of Linux and Android-based IoT platforms, smartphones, and edge devices.
	
	\smallskip
	
	
	This challenge is a part of the \textit{MAI 2026 Workshop and Challenges} consisting of the following competitions:
	
	
	
	\begin{itemize}
		\item 4K Image Super-Resolution on Mobile NPUs~\cite{ignatov2025sr4k}
		\item Real-Time Video Super-Resolution on Smartphone GPUs~\cite{ignatov2025srvideo}
		\item Efficient Image Denoising on Smartphone GPUs~\cite{ignatov20254denoising}
		\item RGB Photo Enhancement on Smartphone GPUs~\cite{ignatov2025rgb}
		\item Learned Smartphone ISP on Mobile GPUs
		\item Bokeh Effect Rendering on Smartphone GPUs~\cite{ignatov2025bokeh}
		\item Efficient LLMs for On-Device Deployment
		\item Adapting Stable Diffusion for Efficient On-Device Inference
	\end{itemize}
	
	\normalsize
	
	
	\noindent The results and solutions obtained in the previous \textit{Mobile AI Challenges} are described in the following papers:
	
	
	\begin{itemize}
		\item Image Super-Resolution on Mobile NPUs~\cite{ignatov2021real,ignatov2022efficient,ignatov2025MAISR}
		\item Learned ISP on Mobile NPUs and GPUs~\cite{ignatov2021learned,ignatov2022learned,ignatov2025MAIRAW}
		\item Video Super-Resolution on Mobile GPUs~\cite{romero2021real,ignatov2022power}
		\item Image Denoising on Mobile GPUs~\cite{ignatov2021fastDenoising}
		\item Bokeh Effect Rendering on Mobile GPUs~\cite{ignatov2022realistic}
		\item Camera Scene Detection on Smartphones~\cite{ignatov2021fastSceneDetection,pouget2021fast}
		\item RGB Photo Enhancement on Mobile GPUs~\cite{ignatov2025MAIRGB}
		\item Depth Estimation on Mobile Devices~\cite{ignatov2021fastDepth,ignatov2022efficient}
	\end{itemize}
	
	\normalsize
	
	\begin{figure*}[t!]
		\centering
		\setlength{\tabcolsep}{1pt}
		\resizebox{0.96\linewidth}{!}
		{
			\includegraphics[width=1.0\linewidth]{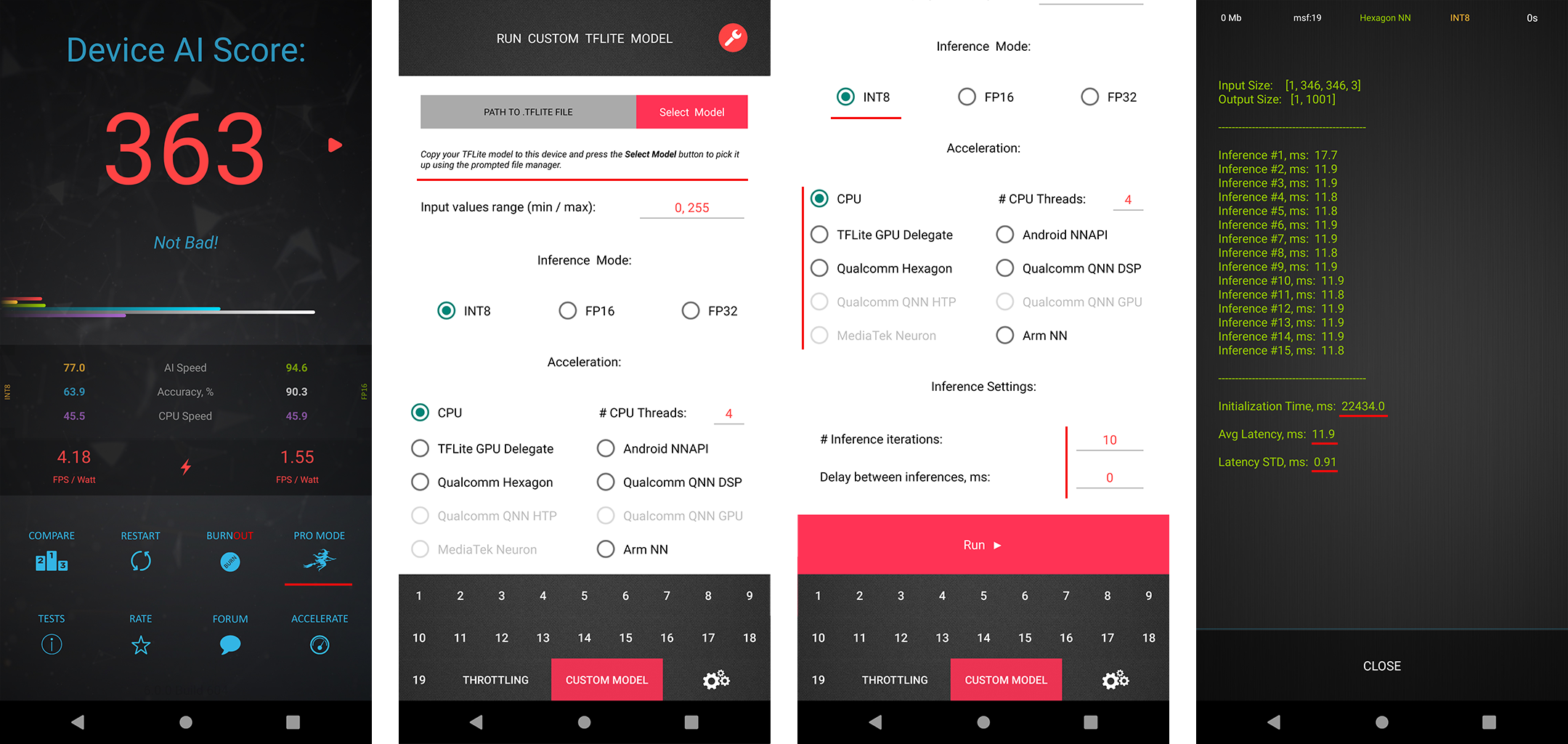}
		}
		\vspace{0.2cm}
		\caption{Loading and running custom TensorFlow Lite models with AI Benchmark application. The currently supported acceleration options include Qualcomm QNN, Hexagon NN, MediaTek Neuron, TFLite GPU, Android NNAPI and ARM NN delegates as well as CPU inference (LiteRT / XNNPACK backends). The latest app version can be downloaded at \url{https://ai-benchmark.com/download}}
		\label{fig:ai_benchmark_custom}
	\end{figure*}

	\section{Challenge}
	
	To design an efficient and practical solution for mobile-related tasks, the following key components are essential:
	
	
	\begin{enumerate}
		\item A high-quality and large-scale dataset that can be used to train and evaluate the solution;
		\item An efficient way to check the runtime and debug the model locally without any constraints;
	\end{enumerate}
	
	This challenge addresses the above issues. Real training data, tools, and runtime evaluation options provided to the challenge participants are described in the next sections.
	
	\subsection{Dataset}
	
	In this challenge, we use the Fujifilm UltraISP dataset~\cite{ignatov2022microisp,ignatov2022pynetv2} collected using the Fujifilm GFX100 medium format 102 MP camera capturing the target high-quality images, and a popular Sony IMX586 Quad Bayer mobile camera sensor that can be found in tens of mid-range and high-end mobile devices released during the past years. The Sony sensor was mounted on the MediaTek Dimensity 820 development board, and was capturing both raw and processed (by its built-in ISP system) 12MP images. The Dimensity board was rigidly attached to the Fujifilm camera, and they were shooting photos synchronously to ensure that the image content is identical. The dataset contains over 6 thousand daytime image pairs captured at a wide variety of places with different illumination and weather conditions. An example set of full-resolution photos from the Fujifilm UltraISP dataset is shown in Fig.~\ref{fig:example_photos}.
	As the collected RAW-RGB image pairs were not perfectly aligned, they were initially matched using the state-of-the-art deep learning based dense matching algorithm~\cite{truong2021learning} to extract 256$\times$256 pixel patches from the original photos. It should be mentioned that all alignment operations were performed on Fujifilm RGB images only, therefore RAW photos from the Sony sensor remained unmodified, exhibiting the same values as read from the sensor.
	
		\begin{table*}[t!]
		\centering
		\resizebox{\linewidth}{!}
		{
			\begin{tabular}{l|c|cc|ccc|ccccc|c}
				\hline
				Team \, & \, Author \, & \, Framework \, & \, Model Size, \, & \, PSNR$\uparrow$ \, & \, SSIM$\uparrow$ \, & \, MOS$\uparrow$ \, & \multicolumn{2}{c}{\, Snapdragon 8 Elite Gen 5\,} & \multicolumn{2}{c}{\, Dimensity 9500\,} & Avg. GPU & \, Final Score \\
				& & & \small{KB} & & & & \, \small{GPU, ms $\downarrow$} \, & \, \small{NPU, ms $\downarrow$} \, &\, \small{GPU, ms $\downarrow$} \, & \, \small{NPU, ms $\downarrow$} \, &  \small{Runtime, ms $\downarrow$} & \\
				\hline
				\hline
				MiAlgo & qyx596 & PyTorch / TensorFlow & 17.4 & 24.19 & 0.8587 & 4.5 & 16.6 & 44.1 & 45.4 & 20 & 27.5 & 96.9 \\
				AVC2 & zhangyongrui & PyTorch & 17.3 & 24.18 & 0.8582 & 4.5 & 59.4 & 40 & 111 & \small{Failed} & 81.2 & 32.4 \\
				UAIC\_ISP & AndreiArhire & PyTorch & 242 & 24 & 0.8625 & 4.8 & 727 & \small{Failed} & 1032 & \small{Failed} & 866.2 & 3.6 \\
				NAF & destiny & TensorFlow & 75.3 & 23.69 & 0.854 & 4.1 & 227 & 731 & 270 & \small{Failed} & 247.6 & 3.1 \\
				MFI & 1514288815 & PyTorch & 94 & 23.76 & 0.8564 & 4.2 & 893 & 365 & 1356 & \small{Failed} & 1100.4 & 0.88 \\
				amonal & amonal & PyTorch & 28.5 & 22.89 & 0.8295 & 1.8 & 275 & 1762 & 359 & \small{Failed} & 314.2 & 0.03 \\
				
				\hline
			\end{tabular}
		}
		\vspace{2.6mm}
		\caption{\small{Mobile AI 2026 Learned Smartphone ISP challenge results and final rankings. During the runtime measurements, the models were processing photos of resolution 1920$\times$2560 pixels on the Snapdragon 8 Elite Gen 5 and Dimensity 9500 SoCs. Team \textit{MiAlgo} is the challenge winner.}}
		\label{tab:results}
	\end{table*}
	
	\subsection{Local Runtime Evaluation}
	
	When developing AI solutions for mobile devices, it is vital to be able to test the designed models and debug all emerging issues locally on available devices. For this, the participants were provided with the \textit{AI Benchmark} application~\cite{ignatov2018ai,ignatov2019ai} that allows to load any custom TensorFlow Lite model and run it on any Android device with all supported acceleration options. This tool contains the latest versions of the \textit{Qualcomm QNN, MediaTek Neuron, Hexagon NN, TFLite GPU, Android NNAPI, Samsung ENN, Amlogic AML} and \textit{Arm NN} delegates, therefore supporting all current mobile platforms and providing the users with the ability to execute neural networks on smartphone NPUs, APUs, DSPs, GPUs and CPUs.
	
	\smallskip
	
	To load and run a custom TensorFlow Lite (LiteRT) model, one needs to follow the next steps:
	
	\begin{enumerate}
		\setlength\itemsep{0mm}
		\item Download AI Benchmark from the official website\footnote{\url{https://ai-benchmark.com/download}} or from Google Play\footnote{\url{https://play.google.com/store/apps/details?id=org.benchmark.demo}} and run its standard tests.
		\item After the end of the tests, enter the \textit{PRO Mode} and select the \textit{Custom Model} tab there.
		\item Put the exported TFLite (LiteRT) model into the \textit{Download} folder of the device.
		\item Select mode type \textit{(INT8, FP16, or FP32)}, the desired acceleration / inference options and run the model.
	\end{enumerate}
	
	\noindent These steps are also illustrated in Fig.~\ref{fig:ai_benchmark_custom}. The instructions on how to generate TFLite model from PyTorch, TensorFlow and Keras pre-trained models were provided in the MAI Workshop GitHub repository~\footnote{\url{https://github.com/aiff22/MAI-2025-Workshop}}.
	
	\subsection{Runtime Evaluation on the Target Platform}
	
	In this challenge, we use two flagship mobile SoCs as our target evaluation platforms: \textit{Qualcomm Snapdragon 8 Elite Gen 5} and \textit{MediaTek Dimensity 9500} chipsets with the \textit{Adreno 840} and \textit{Arm Mali G1-Ultra MC12} GPUs, respectively, that can be found in numerous flagship smartphones. The models were parsed and accelerated using the TFLite GPU delegate~\cite{lee2019device} that demonstrated the best runtime results on both platforms. NPU runtimes provided for reference were obtained by running the models with the Qualcomm QNN and MediaTek Neuron TFLite delegates on the Snapdragon 8 Elite Gen 5 and the Dimensity 9500 SoCs. All final solutions were tested using the aforementioned AI Benchmark application.
	
	\subsection{Challenge Phases}
	
	The challenge consisted of the following phases:
	
	\vspace{-0.8mm}
	\begin{enumerate}
		\item[I.] \textit{Development:} the participants get access to the data and AI Benchmark app, and are able to train the models and evaluate their runtime locally;
		\item[II.] \textit{Validation:} the participants can upload their models to the remote server to check the fidelity scores on the validation dataset, and to compare their results on the validation leaderboard;
		\item[III.] \textit{Testing:} the participants submit their final results, codes, TensorFlow Lite models, and factsheets.
	\end{enumerate}
	\vspace{-0.8mm}
	
	\subsection{Scoring System}
	
	All solutions were evaluated using the following metrics:
	
	\begin{itemize}
		\setlength\itemsep{-0.2mm}
		\item Peak Signal-to-Noise Ratio (PSNR) measuring fidelity score,
		\item Structural Similarity Index Measure (SSIM) as proxy for perceptual score,
		\item Mean Opinion Score (MOS) obtained from a user study and used to evaluate the perceived aesthetic quality of the reconstructed RGB image,
		\item The runtime on the target Snapdragon 8 Elite Gen 5 and Dimensity 9500 GPU.
	\end{itemize}
	
	The score of each final submission was evaluated based on the next formula ($C$ is a constant normalization factor):
	
	\smallskip
	\begin{equation*}
		\text{Final Score} \,=\, \frac{2^{2 \cdot \text{(PSNR + MOS)}}}{C \cdot \text{runtime}},
	\end{equation*}
	\smallskip
	
	During the final challenge phase, the participants did not have access to the test dataset. Instead, they had to submit their final TensorFlow Lite models that were subsequently used by the challenge organizers to check both the runtime and the fidelity results of each submission under identical conditions. This approach solved all the issues related to model overfitting, reproducibility of the results, and consistency of the obtained runtime / accuracy values.
	
	\section{Challenge Results}
	
	In this challenge, 6 teams entered the final phase and submitted their results, TFLite models, codes, executables and factsheets. Table~\ref{tab:results} summarizes the final challenge results for both tracks and report PSNR, SSIM and runtime numbers for the valid solutions on the final test dataset and on the target evaluation platform. The proposed methods are described in section~\ref{sec:solutions}, and the team members and affiliations are listed in Appendix~\ref{sec:apd:team}.
	\subsection{Results and Discussion}

	\begin{figure*}[h!]
		\centering
		\setlength{\tabcolsep}{1pt}
		\resizebox{0.9\linewidth}{!}
		{
			\includegraphics[width=1.0\linewidth]{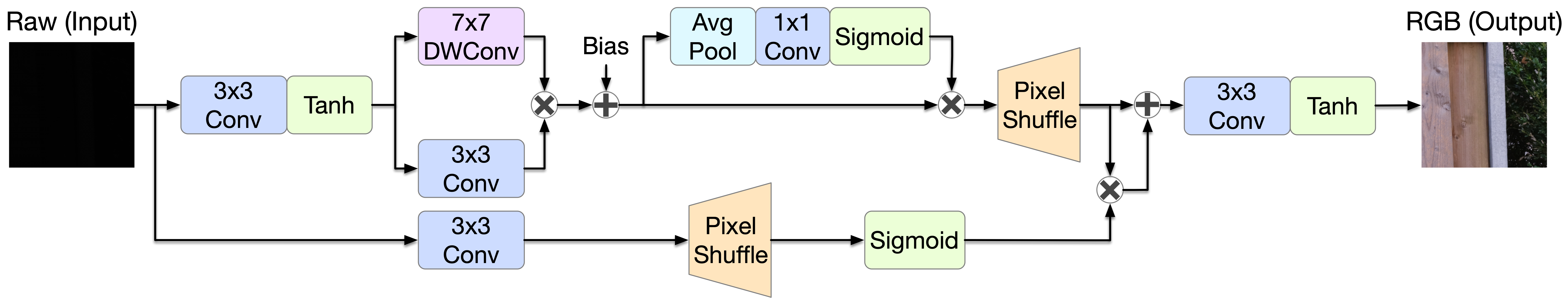}
		}
		\vspace{0.2cm}
		\caption{LKSNet architecture proposed by team MiAlgo}
		\label{fig:MiAlgo}
	\end{figure*}
		
	All submitted solutions demonstrated a very high efficiency: the majority of models are able to process one 5MP (1920$\times$2560 px) image under 0.5 seconds on both target GPUs. This shows great progress in both model design as well as in mobile AI acceleration hardware that is currently able to provide decent speed-up for many common network architectures. Team \textit{MiAlgo} is the challenge winner: the solution proposed by this team was able to demonstrate over 60 and 22 FPS when running on the Adreno and Arm GPUs, respectively. This highlights that the considered model can be potentially applied even for FullHD RAW video processing while achieving a frame rate of over 30 FPS on current mid-range and high-end smartphone SoCs. It should be mentioned that this model is additionally compatible with NPUs from both Qualcomm and MediaTek, yielding an additional 2$\times$ runtime speed-up in the latter case. Very similar numerical and visual results were demonstrated by team \textit{AVC2}, which solution was slightly slower on both platforms. The best perceptual quality was achieved with solution developed by team \textit{UAIC\_ISP}. Despite using a relatively large model, its runtime was still not exceeding one second on the considered mobile GPUs, making it possible to use this solution for practical applications.
	
	A notable progress that should be highlighted is the ease of model conversion to the target TFLite format. Table~\ref{tab:results} shows that the majority of solutions in this challenge were developed using the PyTorch ML framework. Due to the recently presented \textit{litert-torch} plugin~\footnote{\url{https://github.com/google-ai-edge/litert-torch}} developed by the LiteRT team, one can now directly convert the majority of PyTorch models to the TFLite format without the need for TensorFlow / Keras model re-implementation or use of ONNX as an intermediary format. This significantly simplifies the development process of mobile-friendly deep learning architectures for the majority of researchers using PyTorch as their main framework for AI model development and training.
	
	\section{Challenge Methods}
	\label{sec:solutions}
	
	\noindent This section describes solutions submitted by top teams participating in the final stage of the MAI 2026 Learned ISP challenge.
	
	\subsection{MiAlgo}
	
	Team MiAlgo based its solution on the SYENet~\cite{gou2023syenet} architecture. The proposed LKSNet (Large Kernel Skip Net) model (Fig.~\ref{fig:MiAlgo}) starts with 3$\times$3 convolution used to preprocess raw input data. Then, 7$\times$7 depthwise convolution and 3$\times$3 convolution extract features separately and multiply them together. After adding learnable bias and applying channel attention, the features are upsampled by 2$\times$ and multiplied with skip-connected 3$\times$3 convolution features. Finally, the RGB output is obtained after adjustment through 3$\times$3 convolution.
	
	In addition to the SYENet loss function, MSE losses are computed separately for low-frequency and high-frequency image information through FFT-based filtering. The authors also incorporated color loss~\cite{ignatov2017dslr}, LPIPS, SSIM, and cosine similarity losses to better preserve perceptual quality, structural consistency, and feature similarity in the reconstructed images. The entire training split of the dataset was utilized for tuning model parameters with a batch size of 32 and a learning rate of 1e--3.

	\subsection{AVC2}
	
	\begin{figure*}[h!]
		\centering
		\setlength{\tabcolsep}{1pt}
		\resizebox{0.9\linewidth}{!}
		{
			\includegraphics[width=1.0\linewidth]{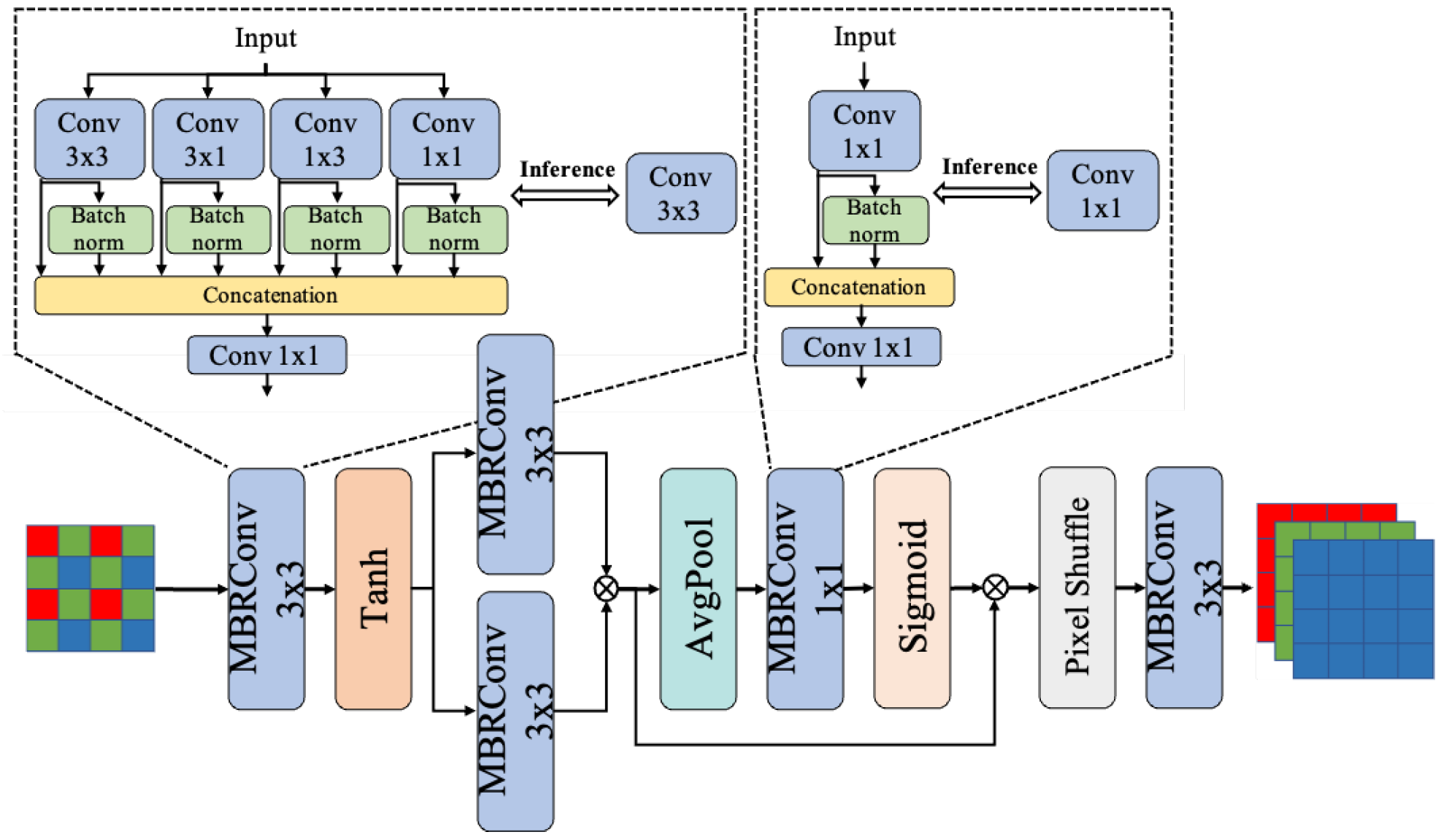}
		}
		\vspace{0.2cm}
		\caption{Network architecture proposed by team AVC2}
		\label{fig:AVC2}
	\end{figure*}
	
	Team AVC2 introduced a simplified version of MobileIE~\cite{yan2025mobileie}, a lightweight image signal processing network designed for mobile deployment that balances performance and efficiency using reparameterization, as illustrated in Figure~\ref{fig:AVC2}.
	The architecture first employs an \textit{MBRConv} $3 \times 3$ layer followed by a \textit{Tanh} activation for initial feature extraction.
	Subsequently, the feature maps are fed into two parallel branches.
	Each branch consists of an \textit{MBRConv} $3 \times 3$ layer with a residual connection.
	The outputs of the two branches are fused via element-wise multiplication.
	After these blocks, a simplified version of the \textit{Hierarchical Dual-Path Attention} (HDPA) module is employed to refine features.
	Upsampling is achieved through a \textit{Pixel Shuffle} layer, followed by a final \textit{MBRConv} $3 \times 3$ layer generating the output image.
	A key aspect is the \textit{MBRConv} block, which utilizes multiple parallel convolutional branches with batch normalization during training; these blocks are reparameterized into standard single convolutional layers for efficient inference.
	
	The network was trained using the $L_1$ loss function, favored for promoting sharper image reconstruction~\cite{bai2025towards,yan2025igdnet,yan2026revisiting}.
	The Adam optimizer was utilized with a cosine annealing learning rate schedule.
	The schedule started with an initial learning rate of 1e--3, preceded by a 10-epoch warm-up phase at a fixed rate of 1e--6.
	The learning rate was subsequently reset every 50 epochs before decaying. The model was trained for a total of 5000 epochs with a batch size of 32.
	The training employed 24000 aligned raw-sRGB image pairs, partitioned into 22000 for training and 2000 for testing.
		
	\subsection{UAIC\_ISP}
	
	Team UAIC\_ISP proposed an improved version of its solutions from the previous learned ISP challenge~\cite{ignatov2025MAIRAW}. The authors made changes to both the architecture and the training method, which resulted in a PSNR score on the validation data that was improved by more than 0.6 dB.
	
\begin{figure*}[h!]
		\centering
		\setlength{\tabcolsep}{1pt}
		\resizebox{0.9\linewidth}{!}
		{
			\includegraphics[width=1.0\linewidth]{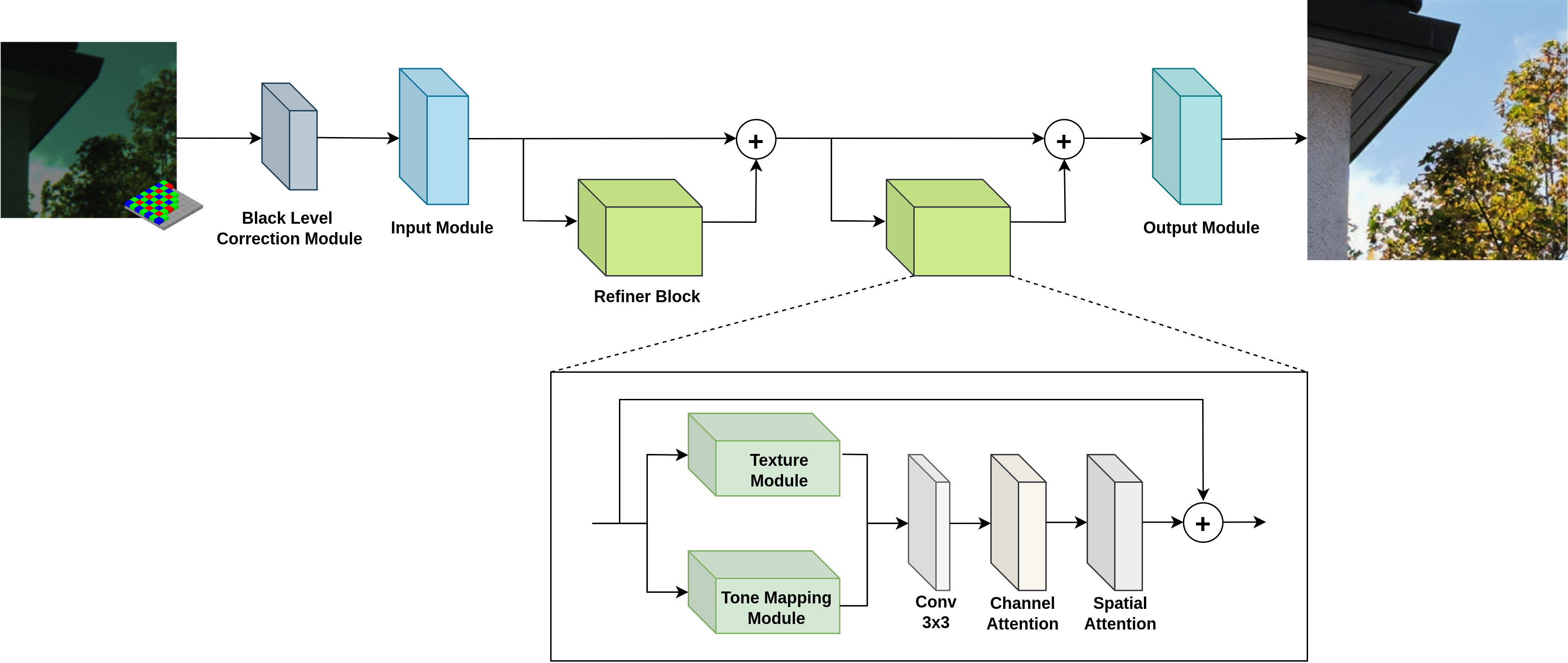}
		}
		\vspace{0.2cm}
		\caption{Deep learning architecture proposed by team UAIC\_ISP}
		\label{fig:UAIC_ISP}
	\end{figure*}

	The proposed architecture is a modified version of RMFA-Net~\cite{li2024rmfa}, in particular the version designed for smartphone inference (Fig.~\ref{fig:UAIC_ISP}). The signal passes through a black level correction module, an input module, two refinement blocks, and one output module. In the first module, the input has its black levels corrected by subtracting learnable per-channel black level offsets, then scaled by learnable per-channel multipliers, and clamped to the valid range [0,1]. In the input module, the input is processed through two 3$\times$3 convolutional layers to extract features and then passed through a \textit{Tanh} activation to map the output to the range (-1, 1). Each block addresses feature representation by incorporating a dedicated texture module, a tone mapping module, and both channel and spatial attention mechanisms. In the texture module, the input is processed through two parallel branches --- a 1$\times$1 convolution for high-frequency details and a 3$\times$3 convolution for low-frequency patterns, then their outputs are concatenated. The tone mapping module processes the input through hierarchical downsampling and upsampling to estimate global illumination features, which are subtracted from the input to produce reflectance according to the Retinex theory~\cite{land1971lightness}. The channel attention module uses global average pooling and 1$\times$1 convolutions to compute channel-wise weights, enhancing important features via sigmoid scaling, while the spatial attention module applies average pooling across channels followed by a 7$\times$7 convolution to generate spatial attention maps that refine feature locations. In the last module, the input is upsampled using pixel shuffle, then refined with a final convolution, and mapped to the [0,1] range with a \textit{Sigmoid} activation.

	The model was trained in four stages with adversarial losses from~\cite{arhire2025learned}. In the first stage, it was trained for $10^{5}$ steps using the Adam optimizer with a fixed learning rate of $5 \cdot 10^{-4}$ and a batch size of 16. Both discriminators had a fixed learning rate of $10^{-5}$ and were trained with Adam at a frequency of 1:10 relative to the neural ISP. The losses and weighting system were the same as in~\cite{arhire2025learned}. The training was then continued for $10^{5}$ more steps. At this time, the color loss was changed so that instead of minimizing the pixel differences in blurred versions,  the Mean Squared Error (MSE) was directly applied. This MSE loss component was weighted 10 times more than any other and the batch size was reduced to 8, while all other details remained the same as in the first stage. During the third stage, the only difference was that the learning rate was decreased from $5 \cdot 10^{-4}$ to $10^{-5}$ by a linear scheduler in the first $10^{5}$ steps. In the last stage, the model was trained for  $2 \cdot 10^{5}$ steps, while the content, total variation loss, and DISTS~\cite{ding2020image} loss were removed, meaning that the only losses used were MSE, LPIPS+~\cite{chen2024topiq} and the adversarial texture losses that rely on the LPIPS+ features. The authors additionally used the manually filtered selection of the dataset, where 17.6\% of the training dataset was removed.
	
	\subsection{NAF}
	
	\begin{figure}[h!]
	\centering
	\setlength{\tabcolsep}{1pt}
	\resizebox{1.0\linewidth}{!}
	{
		\includegraphics[width=1.0\linewidth]{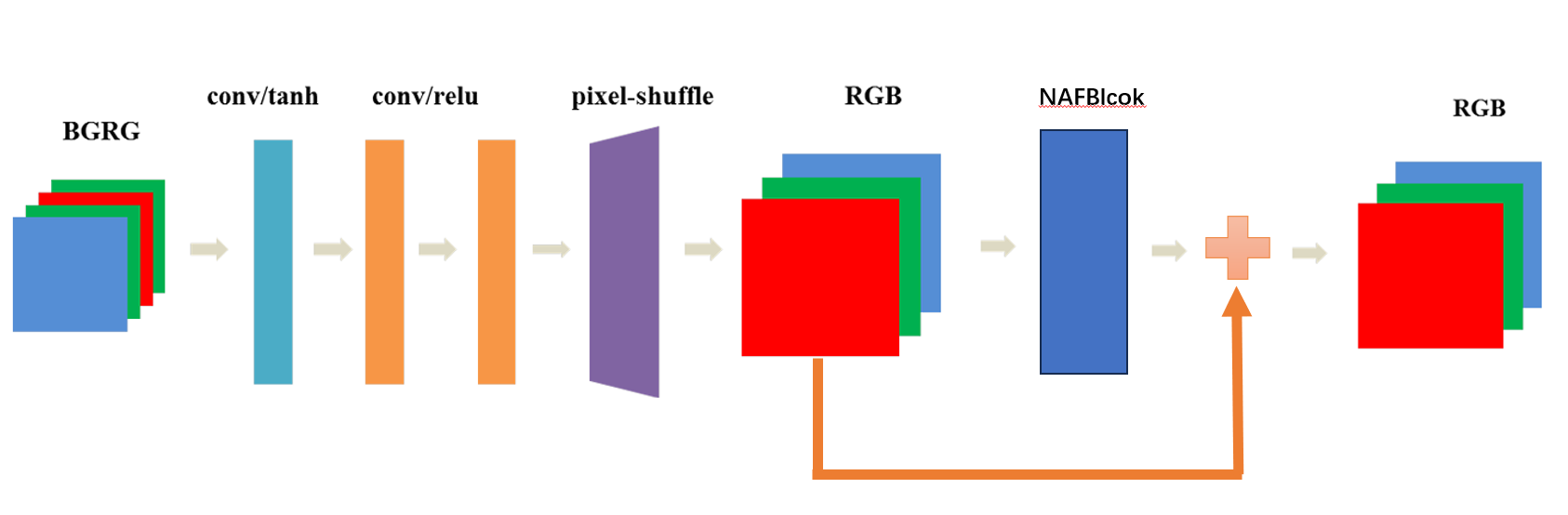}
	}
	\resizebox{0.2\linewidth}{!}{
		\includegraphics[width=0.2\linewidth]{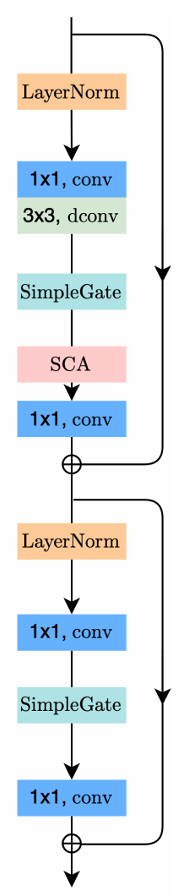}}
	\vspace{0.2cm}
	\caption{NAF-Enhanced Smallnet model (top) and the NAFBlock architecture (bottom) proposed by team NAF}
	\label{fig:NAF}
	\end{figure}
	
	Team NAF proposed an architecture illustrated in Fig.~\ref{fig:NAF}. The model starts with a 4-channel input, first generating basic RGB features through three convolutional layers, dimensional transformation, and color correction (gamma correction and s-curve mapping). Subsequently, it expands the 3-channel RGB features to 32 channels via an upsampling convolution, enhances these features using a \textit{NAFBlock} (that includes \textit{depthwise convolution}, \textit{SimpleGate}, \textit{SCA attention}, and \textit{FFN structure}), reduces the dimensionality back to 3 channels, and combines the result with the original basic RGB features through a residual connection. Finally, it outputs an RGB image within the [0,1] range. The main novelty of the proposed model lies in the integration of a \textit{NAFBlock} to enhance feature representation while preserving the integrity of the original features and improving fine details.
	The model was trained with a combination of the structural similarity loss and the mean absolute error (MAE) loss, balancing pixel-level error minimization and preservation of image structural consistency:
	\begin{align*}
	\mathcal{L}(y_{\text{true}}, y_{\text{pred}}) = 0.15 \cdot \frac{1}{N} \sum_{i=1}^{N} \left(1 - \text{SSIM}(y_{\text{true},i}, y_{\text{pred},i})\right) + \\
	 + \frac{1}{N} \sum_{i=1}^{N} \left| y_{\text{true},i} - y_{\text{pred},i} \right|
	 \end{align*}
	 
	 \subsection{MFI}
	 
	 \begin{figure}[h!]
	 	\centering
	 	\setlength{\tabcolsep}{1pt}
	 	\resizebox{1.0\linewidth}{!}
	 	{
	 		\includegraphics[width=1.0\linewidth]{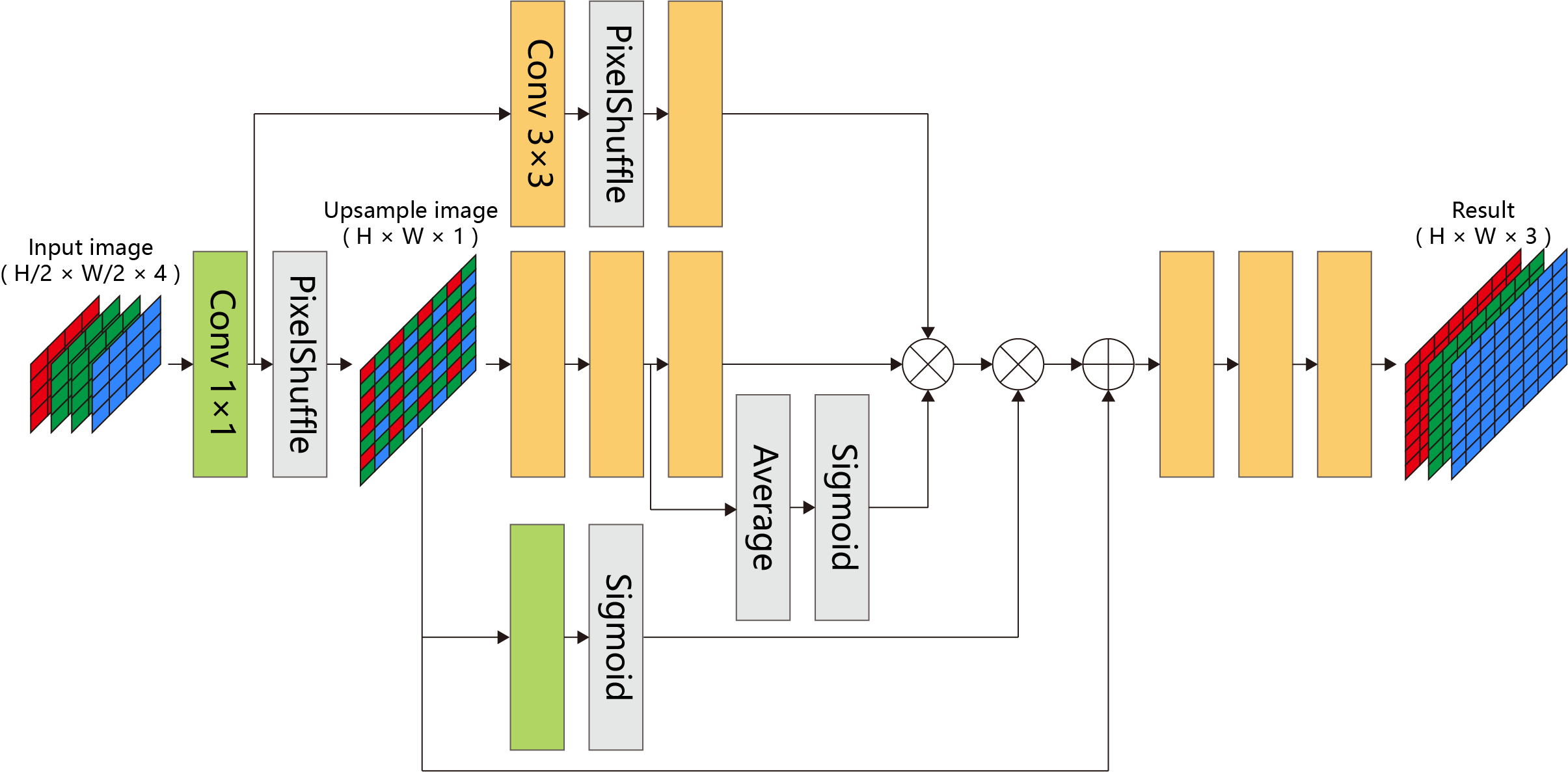}
	 	}
	 	\vspace{0.2cm}
	 	\caption{Multi-level feature interaction model presented by team MFI}
	 	\label{fig:MFI}
	 \end{figure}
	 
	 Team MFI proposed a multi-level feature interaction (MFI) module which architecture is illustrated in Figure~\ref{fig:MFI}. The input of MFI is a 4 channel raw image, which is first processed by a 1 $\times$ 1 convolution layer. Then, MFI splits into two branches for further processing. In the main branch, the feature map is processed by the \textit{PixelShuffle} layer, yielding the upsampled feature map $I_u$. Then the feature map $I_u$ is processed by several 3 $\times$ 3 convolution layers. Inspired by SENet~\cite{hu2018squeeze}, the authors introduced a channel attention module to enhance the feature representation ability of the model and output the first feature map.
	 In the secondary branch, features are processed sequentially by a 3 $\times$ 3 convolution layer, a \textit{PixelShuffle} layer, and a 3$\times$3 convolution layer to obtain the second feature map. 
	 
	 The main branch uses only a 1$\times$1convolution layer before the \textit{PixelShuffle} layer, which can reduce the computational cost of the model and maximally preserve the high-frequency spatial information from the original image, while the secondary branch focuses on the low-frequency information of the image. Then, the two branches interact with each other by element-wise multiplication, which can further enhance the nonlinear processing capability of the model. The fused features are further multiplied element-wise with another feature map derived from $I_u$ after a 1$\times$1 convolution layer and \textit{Sigmoid} activation function, which aims to enhance the retention of high-frequency spatial information. Finally, the resulted feature map is added to $I_u$ and processed by several 3$\times$3 convolution layers to obtain the final RGB image.
	 
	 The model was trained with the $L_1$ loss function and the Adam optimizer. The initial learning rate was set to 1e--4 and decayed by a factor of 0.9 every 10 epochs. The model was trained for 400 epochs with a batch size of 4.
	
	\section{Additional Literature}
	
	An overview of the past challenges on mobile-related tasks together with the proposed solutions can be found in the following papers:
	
	\begin{itemize}
		\item Image Super-Resolution:\, \cite{ignatov2018pirm,cai2019ntire,timofte2018ntire,lugmayr2020ntire}
		\item Learned End-to-End ISP:\, \cite{ignatov2019aim,ignatov2020aim,wirzberger2022lan,ignatov2022microisp,ignatov2022pynetv2}
		\item RGB Image Enhancement:\, \cite{ignatov2018pirm,ignatov2019ntire,ignatov2017dslr,ignatov2018wespe}
		\item Bokeh Effect Rendering:\, \cite{ignatov2019aimBokeh,ignatov2020aimBokeh,ignatov2020rendering}
		\item Image Denoising:\, \cite{abdelhamed2019ntire,abdelhamed2020ntire,flepp2024real}
	\end{itemize}
	
	\section*{Acknowledgements}
	
	We thank AI Witchlabs, ETH Z\"urich, and the University of W\"urzburg (Computer Vision Lab), the organizers and sponsors of this Mobile AI 2026 challenge.
	
	\appendix
	\section{Teams and Affiliations}
	\label{sec:apd:team}
	
	\bigskip
	
	\subsection*{Mobile AI 2026 Team}
	\noindent\textit{\textbf{Title: }}\\ Mobile AI 2026 Learned Smartphone ISP Challenge\\
	\noindent\textit{\textbf{Members:}}\\ Andrey Ignatov$^{1,3}$ \textit{(andrey@vision.ee.ethz.ch)}, Georgy Perevozchikov$^2$ \textit{(georgii.perevozchikov@uni-wuerzburg. de)}, Radu Timofte$^{2,3}$ \textit{(radu.timofte@uni-wuerzburg.de)}\\
	\noindent\textit{\textbf{Affiliations: }}\\
	$^1$ Computer Vision Lab, ETH Zurich, Switzerland\\
	$^2$ Computer Vision Lab, University of W\"urzburg, Germany\\
	$^3$ AI Witchlabs, Switzerland\\
	
	\subsection*{MiAlgo}
	\noindent\textit{\textbf{Members: }}\\ Yuhong He \textit{(1201004027@cnu.edu.cn)}, Yuxuan Qiu, Tianyu Hao, Ruoqi Li, Yueqi Yang, Chaoyu Feng, Dongqing Zou, Lei Lei, Na Jiang\\
	\noindent\textit{\textbf{Affiliations: }}\\
	Xiaomi Inc.\\
	
	\subsection*{AVC2}
	\noindent\textit{\textbf{Members: }}\\ Yongrui Zhang \textit{(zhangyongrui@std.uestc.edu.cn)}, Hailong Yan, Xiangtao Zhang, Ao Li, Le Zhang\\
	\noindent\textit{\textbf{Affiliations: }}\\
	University of Electronic Science and Technology of China\\
	
	\subsection*{UAIC\_ISP}
	\noindent\textit{\textbf{Members: }}\\ Andrei Arhire \textit{(andreiarhire2708@gmail.com)}, Radu Timofte \\
	\noindent\textit{\textbf{Affiliations: }}\\
	Faculty of Computer Science, Alexandru Ioan Cuza University of Iasi (UAIC), Romania\\
		
	\subsection*{NAF}
	\noindent\textit{\textbf{Title:}}\\ NAF-Enhanced Smallnet \\
	\noindent\textit{\textbf{Members: }}\\ Jing Fang \textit{(2627729817@qq.com)}, XingLin Xie, YiChen Lu \\
	\noindent\textit{\textbf{Affiliations: }}\\
	Xidian University, China\\
	
	\subsection*{MFI}
	\noindent\textit{\textbf{Title:}}\\ Multi-level feature interaction for mobile ISP \\
	\noindent\textit{\textbf{Members: }}\\ Haibo Xu \textit{(haibo.xu@zju.edu.cn)} \\
	\noindent\textit{\textbf{Affiliations: }}\\
	College of Optical Science and Engineering, Zhejiang University, China \\
	
	{\small
		\bibliographystyle{ieeenat_fullname}
		\bibliography{egbib}
	}
	
\end{document}